\documentclass[runningheads]{llncs}
\usepackage[T1]{fontenc}
\usepackage{graphicx}
\usepackage{amsmath,amssymb,amsfonts}
\usepackage{relsize}
\usepackage{multirow}
\usepackage{booktabs}
\usepackage{comment}
\usepackage{cite}
\usepackage{enumitem}
\usepackage{caption}
\usepackage{xcolor}
\usepackage{nicefrac}

\newcommand{\slicemodule}{$\texttt{Slice\;Module}$}
\newcommand{\srmodule}{\texttt{SR\;Module}}

\newcommand{\TE}{\text{TE}}

\begin{document}
\title{Physics-Informed Joint Multi-TE Super-Resolution with Implicit Neural Representation for Robust Fetal $\text{T2}$ Mapping}
\titlerunning{Physics-Informed Joint Multi-TE SR for $\text{T2}$ Mapping}
%

\author{
  Busra Bulut\inst{1,2}\textsuperscript{*} \and
  Maik Dannecker\inst{3}\textsuperscript{*}\and
  Thomas Sanchez\inst{1,2}\and
  Sara Neves Silva\inst{6}\and
  Vladyslav Zalevskyi \inst{1,2}\and
  Steven Jia \inst{7} \and
  Jean-Baptiste Ledoux\inst{1,2} \and
  Guillaume Auzias \inst{7} \and
  François Rousseau\inst{8} \and
  Jana Hutter \inst{4,9}\and
  Daniel Rueckert\inst{3,4,5}\and
  Meritxell Bach Cuadra\inst{1,2}}
%

\authorrunning{B. Bulut, M. Dannecker et al.}
%

\institute{Department of Radiology, Lausanne University Hospital and University of Lausanne, Lausanne, Switzerland \and CIBM Center for Biomedical Imaging, Lausanne, Switzerland\and 
Chair for AI in Healthcare and Medicine, Technical University of Munich (TUM) and TUM University Hospital, Munich, Germany
\and
Department of Computing, Imperial College London, UK
\and 
Munich Center for Machine Learning (MCML), Munich, Germany
\and
Biomedical Engineering Department,School of Biomedical Engineering and Imaging Sciences, King’s College London, London, UK
\and 
Institut de Neurosciences de la Timone, UMR 7289, CNRS, Aix-Marseille Université, 13005, Marseille, France
\and 
IMT Atlantique, LaTIM UMR 1101, Brest, France
\and 
Smart Imaging Lab, Friedrich-Alexander University Erlangen, Germany\\[2mm]
\textsuperscript{*}These authors contributed equally to this work.\\
\email{busra.bulut@unil.ch}}


%
\maketitle              
\looseness=-1
\begin{abstract}$\text{T2}$ mapping in fetal brain MRI has the potential to improve characterization of the developing brain, especially at mid-field (0.55T), where T2 decay is slower. However, this is challenging as fetal MRI acquisition relies on multiple motion-corrupted stacks of thick slices, requiring slice-to-volume reconstruction (SVR) to estimate a high-resolution (HR) 3D volume. 
Currently, T2 mapping involves repeated acquisitions of these stacks at each echo time (TE), leading to long scan times and high sensitivity to motion. 
We tackle this challenge with a method that jointly reconstructs data across TEs, addressing severe motion. Our approach combines implicit neural representations with a physics-informed regularization that models T2 decay, enabling information sharing across TEs while preserving anatomical and quantitative T2 fidelity. We demonstrate state-of-the-art performance on simulated fetal brain and \textit{in vivo} adult datasets with fetal-like motion. We also present the first \textit{in vivo} fetal T2 mapping results at 0.55T. Our study shows potential for reducing the number of stacks per TE in T2 mapping by leveraging anatomical redundancy.

\keywords{Quantitative mapping \and Slice-to-volume reconstruction \and Implicit Neural Representation}
\end{abstract}
\vspace{-.4cm}
\section{Introduction}
\vspace{-.2cm}

Magnetic Resonance Imaging (MRI) is a key tool in perinatal neuroimaging due to its non-invasive nature and excellent soft tissue contrast \cite{Rutherford2009}. T2-weighted (T2w) imaging is especially valuable for assessing tissue maturation and detect brain abnormalities in early brain development \cite{Hassink1992} but scanner and protocol variability limit its use in multicenter studies \cite{Schmidbauer2021_v2}. 
Quantitative T2 mapping offers consistent, field-strength–dependent voxel-wise T2 relaxation times, improving the detection of pathologies through comparison with normative data \cite{Schmidbauer2021, verdera2025real}. At 0.55T, T2 mapping benefits from slower T2 decay and a broader dynamic range, enhancing visualization of subtle tissue contrasts, especially in low-signal areas such as deep gray matter \cite{uus2023combined,aviles2023reliability,verdera2025real}. Its low cost makes it a promising MRI option for low-resource settings, supporting standardized brain maturation assessment in diverse, high-risk populations \cite{Arnold2022}.


 To deal with unexpected fetal motion and preserve signal-to-noise ratio (SNR), single-shot fast spin echo (SS-FSE) acquisitions are used to acquire 2D stacks of thick T2w slices, effectively freezing \emph{intra}-slice fetal motion in time~\cite{uus2023retrospective}. \emph{Inter}-slice motion and anisotropic resolution remain problematic, necessitating post-acquisition slice-to-volume reconstruction (SVR) techniques to achieve coherent, HR 3D volumes~\cite{Rousseau2006,Jiang2007,Kuklisova2012,Tourbier2015,ebner2020automated}. These methods iteratively estimate slice-to-volume transformations, followed by super-resolution (SR) reconstruction of a HR 3D volume. Traditional grid-based SVR methods are computationally demanding and often fail in the presence of severe motion~\cite{uus2023retrospective,Xu2023}. Recent approaches employing implicit neural representations (INRs) \cite{Mildenhall2021,Sitzmann2020} overcome these limitations by continuously modeling image intensities, enabling resolution-independent reconstruction and improved computational efficiency~\cite{Wu2021,Xu2023,mcginnis2023single,jia2024joint,dannecker2025meta}. However, except for \cite{jia2024joint,mcginnis2023single}, these approaches are mono-modal, thus sub-optimal for multi-contrast setups. In addition, to our knowledge, no existing INR method reliably preserves quantitative information of intensities, making their applicability to quantitative analyses such as T2 mapping largely unexplored.

 Quantitative T2 mapping with SS-FSE requires multiple acquisitions at varying TEs, which exacerbates the impact of fetal motion. Additionally, mid-field (0.55T) imaging introduces lower SNR, complicating accurate reconstruction and T2 quantification. While existing methods can be applied to mid-field MRI, the only method routinely used for this is SVRTK~\cite{Kuklisova2012,Bhattacharya2024,uus2025scanner}. However, while SVRTK can be applied to data acquired at different TEs, it independently reconstructs each TE ignoring the shared anatomical information. Other adult and phantom studies at 0.55T have utilized interpolation-based methods for T2 quantification, achieving reliable T2 estimates but not addressing severe motion artifacts inherent in fetal imaging~\cite{lajous2020t2,margaux}. Finally, studies targeting fetal imaging at 0.55T have primarily focused on T2* mapping using echo-planar imaging (EPI) sequences: this leads to a fast 3D volume acquisition (no SVR required), but comes at the cost of lower resolution and lower anatomical contrast 
~\cite{vasylechko2015t2,Blazejewska2017,verdera2025real,uus2023combined,Payette2024}.  

 Our method addresses these gaps by integrating a joint INR-based reconstruction framework explicitly tailored for robust fetal T2 mapping at mid-field based on SS-FSE acquisitions. Specifically, we propose the first physics-informed joint multi-TE super-resolution reconstruction (SRR) framework using INRs. Our key contributions can be summarized as follows: \textbf{(i)} We jointly reconstruct all TEs within a shared continuous representation to leverage both anatomical 
 redundancy and the exponential signal decay across TEs, achieving state-of-the-art reconstruction quality and robustness to severe motion with signal drop-out stacks. \textbf{(ii)} Our setting is fully self-supervised, enabling robustness to out-of-domain TEs and mid-field settings. \textbf{(iii)} We observe promising results with only 1 or 2 stacks per TE, achieving high-quality reconstruction and T2 mapping, suggesting the potential to reduce acquisition time. \textbf{(iv)} We present the first T2 maps of the fetal brain acquired at 0.55T.

\vspace{-.3cm}
\section{Methodology}
\begin{figure}[t]
    \centering
    \includegraphics[width=\textwidth]{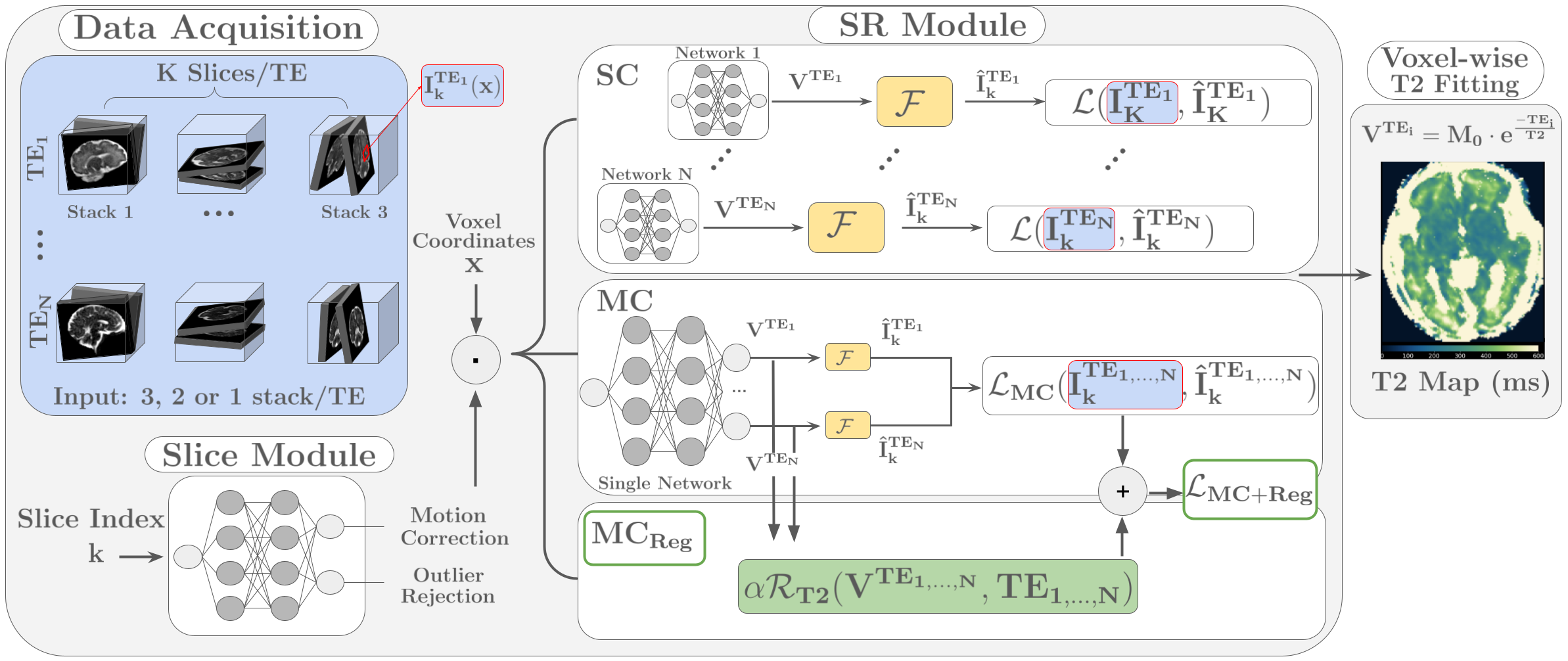}
\caption{Overview of the framework with \slicemodule{} and 3 variants of \srmodule{}: SC, MC, and $\text{MC}_{\mathrm{Reg}}$, our physics-informed approach (in green).}
    \label{fig:mc_reg}
    \vspace{-.5cm}
\end{figure}

\vspace{-.2cm}
We build upon the SIREN-based SVR (SSVR) design by Dannecker et al.~\cite{dannecker2025meta}. 
 We extend the SSVR framework to simultaneously reconstruct the HR volumes acquired at all TEs by leveraging data redundancy across them, while preserving the T2 quantitative metrics, as illustrated in Figure~\ref{fig:mc_reg}. 
SSVR approach uses INR models to tackle both slice-to-volume registration and reconstruction tasks, similarly to NeSVoR~\cite{Xu2023}. But, while NeSVoR relies on slice pre-alignment via the pre-trained transformer model SVoRT~\cite{svort} for severe motion, SSVR is fully self-supervised, relying on two sinusoidal representation networks (SIREN)~\cite{Sitzmann2020}, respectively tackling motion correction with outlier handling and SR reconstruction.
We provide detailed description below. 

\subsection{Problem Setting}
We aim at estimating $N$ unknown HR volumes $V^{\text{TE}_i}$, where $i$ indexes the $N$ different TEs. Each TE comes with 3 stacks of 2D slices. We treat each slice as an independent observation tied only to its echo time (not to its specific stack). For notational clarity and without loss of generality, we assume the total number of slices across all stacks of one TE to be $K$.

\noindent\textbf{\slicemodule{}.} Following SSVR~\cite{dannecker2025meta}, a 3-layer SIREN, operating on slices only, performs motion correction and outlier handling. For each slice $k$, its rigid transformation $\mathbf{T}_{\psi_k}$ is predicted. For a detailed explanation of outliers coefficients, please refer to Dannecker et al.~\cite{dannecker2025meta}.

\noindent\textbf{\srmodule{}.}
Our self-supervised training scheme follows the principles detailed in prior INR-based SRR works \cite{Xu2023,jia2024joint,dannecker2025meta}. It operates voxel-wise: it takes a 3D voxel coordinate $\mathbf{x}$ as input and predicts the corresponding HR intensity $\{V^{TE_i}(\mathbf{x})\}_{i=0}^N$. Then, we simulate low-resolution slices $\{\hat{I}^{TE_i}_{k}(\mathbf{x})\}_{k=1}^K$ by modeling the physical acquisition process (i.e., applying the standard continuous forward model $\mathcal{F}$ including the point-spread function)\cite{Xu2023,dannecker2025meta,jia2024joint}. Next, for each slice $k\in[1 \ldots K]$ we minimize the differences between the real slice measurements $I^{TE_i}_{k}(\mathbf{x})$ and the simulated low-resolution slice $\hat{I}^{TE_i}_{k}(\mathbf{x})$ according to
\begin{equation}
\label{eq:loss_function_sc}
\mathcal{L}(V^{TE_i}) = \frac{1}{K}\sum_{k=1}^{K} \frac{1}{\lvert X \rvert}\sum_{\mathbf{x}\in X} \lvert I^{TE_i}_k(\mathbf{T}_{\psi_k}\mathbf{x})-\hat{I}_k^{TE_i}(\mathbf{x})\lvert,
\end{equation}
where $X$ is the set of coordinates and $\mathbf{T}_{\psi_k}$ the predicted rigid motion correction.


\subsection{T2 Fitting and Physics-Informed Regularization : $R_{T2}$}
\label{subsec:regul}

The evolution of the signal across TEs in a voxel of the T2-w HR volume can be modeled by a monoexponential decay derived from the Bloch equations~\cite{haacke1999magnetic}:
$V^{\TE}(\mathbf{x}) = M_0(\mathbf{x}) \exp\left(-\TE/T_2(\mathbf{x})\right)$,
where $M_0(\mathbf{x})$ is the initial magnetization and $T_2(\mathbf{x})$ is the quantitative metric of interest. Taking the logarithm yields a linear equation $\log(V^{\TE}(\mathbf{x})) = \log(M_0(\mathbf{x})) - \TE/T_2(\mathbf{x})$, that after defining the following variables as: 
$$
y(\textbf{x}) = \begin{bmatrix} \log(V^{\TE_1}(\textbf{x})) \\ \vdots \\ \log(V^{\TE_N}(\textbf{x})) \end{bmatrix}, \quad D = \begin{bmatrix} 1 & \TE_1 \\
\vdots & \vdots\\ 1 & \TE_N \end{bmatrix}, \quad
\beta(\mathbf{x}) = \begin{bmatrix} \log(M_0(\textbf{x})) \\ -1/T_2(\textbf{x})\end{bmatrix},
$$ takes the form of $y(\mathbf{x})= D \beta(\mathbf{x})$. For clarity, we omit the explicit dependency on voxel position $\mathbf{x}$ of $y$ and $\beta$ in the following equations. The parameter vector $\beta$ can then be estimated using ordinary least squares (OLS):
\begin{equation*}
\min_\beta \|D\beta - y\|_2^2 = \|D \beta^{*} - y\|_2^2 , \text{with } \beta^{*} = (D^T D)^{-1} D^Ty
\end{equation*}
We leverage this voxel-wise constraint to regularize the output of our $\text{MC}_{\text{Reg}}$. We encourage the predictions $y$ to minimize the residual of OLS i.e. the deviation from the exponential model. This yields the following regularization term.
\begin{equation*}
r_{T2}(y)= \|D(D^\top D)^{-1}D^\top y - y\|_2^2 = \|A y\|_2^2
\end{equation*}
where $A = D(D^\top D)^{-1}D^\top - I$ is a projection residual matrix not depending on the voxel position $\mathbf{x}$ and can hence be precomputed. Then, our final regularization $R_{T2}$ is computed over all coordinates as:
\begin{equation*}
R_{T2}(V^{TE_1},..,V^{TE_N}) = \frac{1}{\lvert X \rvert}\sum_{\mathbf{x}\in X} r_{T2}(y(\mathbf{x})) ,
\end{equation*}

\noindent This regularization term enforces consistency with the underlying T2 decay. Since INRs output $y$ are inherently computed voxel-wise, this formulation is not only conceptually aligned but also computationally efficient, making voxel-wise regularization particularly suitable in our setting.

\subsection{Model Setup}
We evaluate 3 variants of \srmodule{} (see Figure~\ref{fig:mc_reg}).
\begin{enumerate}[leftmargin=*]
  \item \textit{Single-contrast} (SC)~\cite{dannecker2025meta}: Each TE (\textit{one contrast}) is reconstructed by an independent SIREN~\cite{Sitzmann2020} network minimizing Eq.\ref{eq:loss_function_sc}.
  \item \textit{Multi-contrast} (MC): This corresponds to the setting of Jia et al.~\cite{jia2024joint}, except that we use the SSVR architecture~\cite{dannecker2025meta} instead of NeSVoR~\cite{Xu2023}. A single SIREN~\cite{Sitzmann2020} network jointly predicts all $N$ TEs, as multiple output channels. This yields the loss 
    \begin{equation}
    \label{eq:loss_function_mc}
    \mathcal{L}_\textit{MC} = \mathlarger{\frac{1}{N} \sum_{i=1}^N \frac{1}{K}\sum_{k=1}^{K}} \frac{1}{\lvert X \rvert}\sum_{\mathbf{x}\in X} \lvert I^{TE_i}_k(\mathbf{T}_{\psi_k}\mathbf{x})-\hat{I}_k^{TE_i}(\mathbf{x})\lvert,
    \end{equation}
    \item \textit{Ours,} $\text{MC}_{\mathrm{Reg}}$ : We further extend (\ref{eq:loss_function_mc}) by adding a physics-informed regularization to enforce better exchange of information among TEs and a more consistent representation. We minimize \(\mathcal{L}_{\textit{MC}_\textit{Reg}}=\mathcal{L}_\textit{MC}+\alpha R_{T2}\).
\end{enumerate}
We compare these variants to the state-of-the-art SVR method that enables post-processing for T2 mapping: SVRTK~\cite{Kuklisova2012,uus2025scanner}. Indeed, the current version of NeSVoR~\cite{Xu2023} does not preserve the original intensity ranges across TEs, yielding non-physical values when fitting the exponential decay for T2 mapping. We could then not use it as baseline.

\section{Experiment Setting}

\begin{figure}
\centering
\includegraphics[width=.6\textwidth]{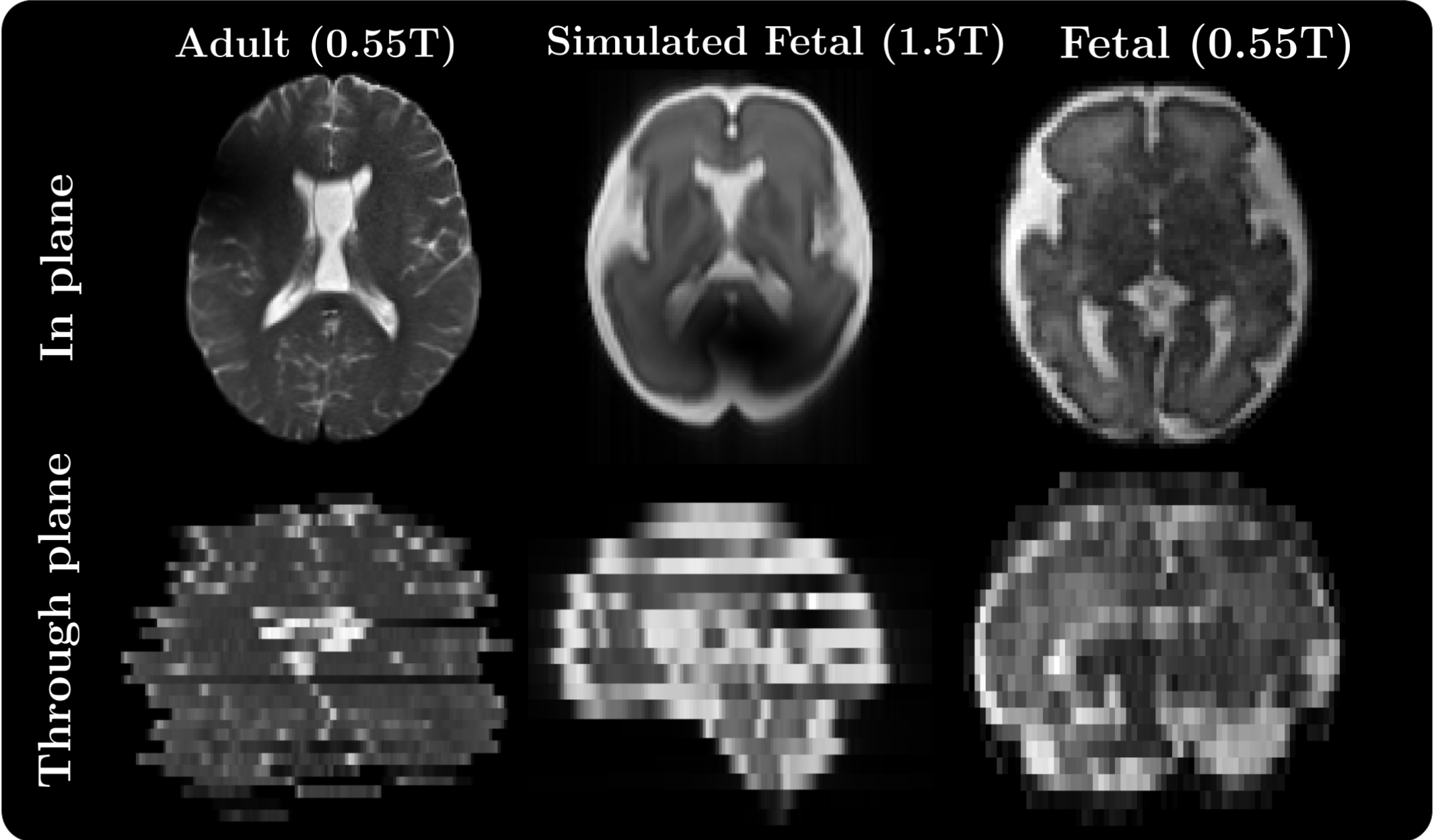}
\vspace{-.1cm}
\captionof{figure}{Example of in-plane and through-plane slices of the different data sets used in the study (resp. TE=114 ms, 220 ms, 300 ms).}\label{fig:data_example}
\end{figure}

\looseness=-1
\noindent\textbf{Data.} We evaluate our method using three datasets described in Table~\ref{tab:data}. 
\begin{enumerate}[leftmargin=*]
    \item \textit{Simulated fetal brain data:} With FaBiAN~\cite{lajous2022fetal,lajous2025dataset}, we simulated motion-corrupted T2w SS-FSE acquisitions at 25, 27, 30, 33, 35 weeks gestational age (GA). Signal dropouts based on \cite{svort} are also simulated (Fig.~\ref{fig:data_example}). A HR motion-free volume at 0.8 $mm^3$ resolution (res.) that can be used to fit reference T2 maps is also simulated.  
    \item In-vivo \textit{adult brain data}\cite{margaux}: 10 volunteers are scanned at 0.55T Siemens MAGNETOM Free.Max (Erlangen, Germany) MRI scanner at Lausanne Hospital (approved by the ethics committee of the Canton of Vaud, Switzerland, project number: 2024-04995). The protocol used SS-FSE sequences to mimic acquisitions for fetuses. Typical fetal MRI artifacts like inter-slice motion and signal dropouts were simulated during post-processing using the method described in [29] (Fig.~\ref{fig:data_example}). Reference SRR and T2 estimation are derived from the initial volumes via resampling to 1 $mm^3$, denoising and trilinear interpolation \cite{margaux}.
    \item In-vivo \textit{fetal brain data:} We acquired data from 2 fetus (GA=32, 37) using SS-FSE sequences 
    at St. Thomas' Hospital, London, under the ethically approved MEERKAT [REC: 21/LO/0742], MiBirth [REC: 23/LO/0685], NANO [REC: 22/YH/0210] and approved for sharing with interested academic researchers around the world by the Ethics Committee London Bromley (Ethics code 21/LO/0742). No reference T2 maps are available for this experiment. Data were denoised prior to reconstruction \cite{tustison_antsx_2021}.
\end{enumerate}
For all cases, three stacks per TE were acquired (axial, coronal, sagittal) for a total of 3 TEs ($N =3$). Note, TE values vary by scenario. For FaBiAN, they match typical 1.5T values \cite{Bhattacharya2024}. In adults, TEs span the expected T2 decay range \cite{margaux}. For in-vivo fetal imaging, TEs were heuristically adapted from 1.5T \cite{Bhattacharya2024}, accounting for longer T2 at 0.55T.

\begin{table}[t]
\centering
\caption{Description of the datasets and the HR reconstruction resolution (res.).}\label{tab:data}
\begin{tabular}{lcccccc}
\toprule
 
\multirow{2}{*}{\textbf{Dataset}} & \multirow{2}{*}{\textbf{Participants}} & \textbf{Resolution} & \textbf{Magnetic} & \textbf{TEs}  & \textbf{TR} & \textbf{HR Res.}\\
& & \textbf{(mm\textsuperscript{3})} &\textbf{Field}& \textbf{(ms)} & \textbf{(ms)}&\textbf{(mm\textsuperscript{3})} \\
\midrule
FaBiAN
  & 5 & 0.8$\times$0.8$\times$4.5 
  & 1.5T & 220, 500, 690 & 12 &0.8$\times$0.8$\times$0.8\\
Adult 
  & 10 & 1.18$\times$1.18$\times$4.5 & 0.55T & 114, 200, 299 &2.5& 1$\times$1$\times$1\\
Fetal
  & 2 & 1.48$\times$1.48$\times$4.5 
  & 0.55T & 300, 397, 600 & 2.5&0.8$\times$0.8$\times$0.8\\
\bottomrule
\end{tabular}
\end{table}

\noindent\textit{Training setting.}
 The INR-based methods were trained for 50 epochs on the adult and simulated fetal data, and for 300 epochs on \textit{in vivo} fetal data until convergence on a 48GB NVIDIA RTX6000 GPU. The rest of the hyperparameters were kept the same as in \cite{dannecker2025meta}. We set $\alpha$, balancing intensity loss and regularization loss, empirically for each dataset. Specifically, we set $\alpha= 0.5$ for adult and simulated fetal data, $\alpha= 10$ for 1 stack/TE settings to compensate for the sparse data, and $\alpha= 1.0$ for \textit{in vivo} fetal experiments due to the poorer image quality than in simulated data. Our code will be made publicly available upon acceptance of the paper \cite{dannecker2025meta}.

\noindent\textit{Post Processing.}
Brain extraction was performed using \cite{https://doi.org/10.48550/arxiv.1901.11341}, followed by rigid co-registration across TEs \cite{tustison_antsx_2021}. T2 maps were segmented using \texttt{SynthSeg} \cite{Billot2023} for adult brains and \texttt{DRIFTS} \cite{zalevskyi2024maximizing} for fetal brains.

\noindent\textbf{Experiments.} We validate our method with three experiments:
\begin{enumerate}[leftmargin=*]
    \item Simulated fetal data: quantitative validation is done on (i) reconstructions through SSIM~\cite{SSIM} to the GT and (ii) T2 mapping with mean absolute error (MAE) of the predicted T2 relaxation times of white matter (WM) and gray matter (GM).
    \item In-vivo adult data: we run three scenarios with 3, 2, and 1 input stack(s) per TE respectively. We compare $\text{MC}_{\mathrm{Reg}}$ over MC and SC \cite{dannecker2025meta} (SVRTK is not adapted for adult data) using the same metrics as for the fetal data, SSIM and MAE. Note, in the 1-stack-per-TE setting, each of the three stacks corresponds to an orthogonal view.
    \item \textit{In-vivo} fetal data at 0.55T: qualitatively evaluation of SVRTK and our proposed framework $\text{MC}_{\mathrm{Reg}}$ is done using either 3 or 2 stacks per TE to evaluate their robustness to sparse input. 

\end{enumerate}
To assess statistical differences between different reconstruction methods, a pairwise two-sample t-test was performed on the regional T2 estimates.

\begin{table}[t]
\centering
\captionof{table}{Quantitative results using the simulated FaBiAN data. Mean Absolute Errors (MAE) of the computed T2 maps and SSIM of HR reconstructions.}


\label{tab:quant_results}
\begin{tabular}{lccccc}
\toprule
\textbf{Model} & \textbf{MAE$_{\text{WM}}\downarrow$} & \textbf{MAE$_{\text{GM}}\downarrow$} & \textbf{SSIM$_{\text{TE}=220}\uparrow$} & \textbf{SSIM$_{\text{TE}=500}\uparrow$} & \textbf{SSIM$_{\text{TE}=690}\uparrow$} \\
\midrule
SVRTK & 39.00 ± 2.10 & 66.20 ± 7.54 & 0.73±0.06 & 0.77±0.02 & 0.71±0.05 \\
SC & 22.79±~1.52 & 39.64±~4.17 & 0.78±0.02 & 0.83±0.02 & 0.83±0.02 \\
MC & 21.48±~1.73 & 33.98±~4.46 & 0.79±0.02 & \textbf{0.84±0.01} & \textbf{0.84±0.02 }\\
$\text{MC}_{\text{reg}}$ & \textbf{20.88±~1.75} & \textbf{32.71±~5.43} & \textbf{0.80±0.02} & \textbf{0.84±0.01} & 0.83±0.03 \\
\bottomrule
\end{tabular}

\centering
\caption{Comparison of performance metrics across reconstruction methods for adult data with variable stacks/TE. SC fails for 1 stack/TE. Best scores in \textbf{bold}.}\label{tab:adult_data_results}
\resizebox{\columnwidth}{!}{%
\begin{tabular}{clccccc}
\toprule
\textbf{\nicefrac{Stacks}{TE}}&\textbf{Model} &\textbf{MAE$_{\text{WM}}\downarrow$} & \textbf{MAE$_{\text{GM}}\downarrow$} & \textbf{SSIM$_{\text{TE}=114}\uparrow$} & \textbf{SSIM$_{\text{TE}=202}\uparrow$} & \textbf{SSIM$_{\text{TE}=299}\uparrow$} \\
\midrule
\multirow{3}{*}{3}
& SC                & $19.22 \pm 7.04$ & $18.66 \pm 6.49$ & $0.77 \pm 0.02$ & $0.77 \pm 0.02$ & $0.69 \pm 0.09$ \\
& MC                & $\mathbf{12.98 \pm 5.33}$ & $18.07 \pm 7.57$ & $\mathbf{0.79 \pm 0.02}$ & $ \mathbf{0.79 \pm 0.02}$ & $ \mathbf{0.77 \pm 0.02}$ \\
& MC$_\mathrm{Reg}$ & $14.88 \pm 4.60$ & $\mathbf{17.65 \pm 6.20}$  & $\mathbf{0.79 \pm 0.02}$ & $ \mathbf{0.79 \pm 0.02}$ & $ \mathbf{0.77 \pm 0.02}$ \\
\midrule
\multirow{3}{*}{2} 
& SC                & $53.44 \pm 12.54$& $59.17 \pm 37.36$& $0.71 \pm 0.06$ & $0.62 \pm 0.09$ & $0.48 \pm 0.11$ \\
& MC                & $ \mathbf{18.36 \pm 9.14}$ & $20.41^* \pm 7.32$ & $0.76 \pm 0.03$ & $0.76 \pm 0.02$ & $ \mathbf{0.73 \pm 0.03}$ \\
& MC$_\mathrm{Reg}$ & $18.44 \pm 8.54$ & $ \mathbf{19.93 \pm 7.00}$ & $ \mathbf{0.77 \pm 0.03}$ & $ \mathbf{0.77 \pm 0.02}$ & $ \mathbf{0.73 \pm 0.03}$ \\
\midrule
\multirow{2}{*}{1}
& MC                & $123.57 \pm 22.67$ & $215.87 \pm 151.83$ & $0.34 \pm 0.04$ & $0.35 \pm 0.04$ & $0.34 \pm 0.04$ \\
& MC$_\mathrm{Reg}$ & $ \mathbf{62.40^* \pm 24.49}$ & $ \mathbf{58.68^* \pm 36.49}$ & $ \mathbf{0.61^* \pm 0.07}$ & $ \mathbf{0.69^* \pm 0.06}$ & $ \mathbf{0.57^* \pm 0.06}$ \\
\bottomrule
  \multicolumn{7}{l}{%
    \scriptsize
    \shortstack[l]{%
      \\[0.05ex] 
      *Value significantly better than the second best ($p<0.05$, paired t‐test).
    }
  }
\end{tabular}}
\end{table}

\section{Results and Discussion}
\subsubsection{Experiment 1 -- Simulated Fetal Data at 1.5T.} 
\begin{figure}[h!]
    \centering
    \includegraphics[width=\textwidth]{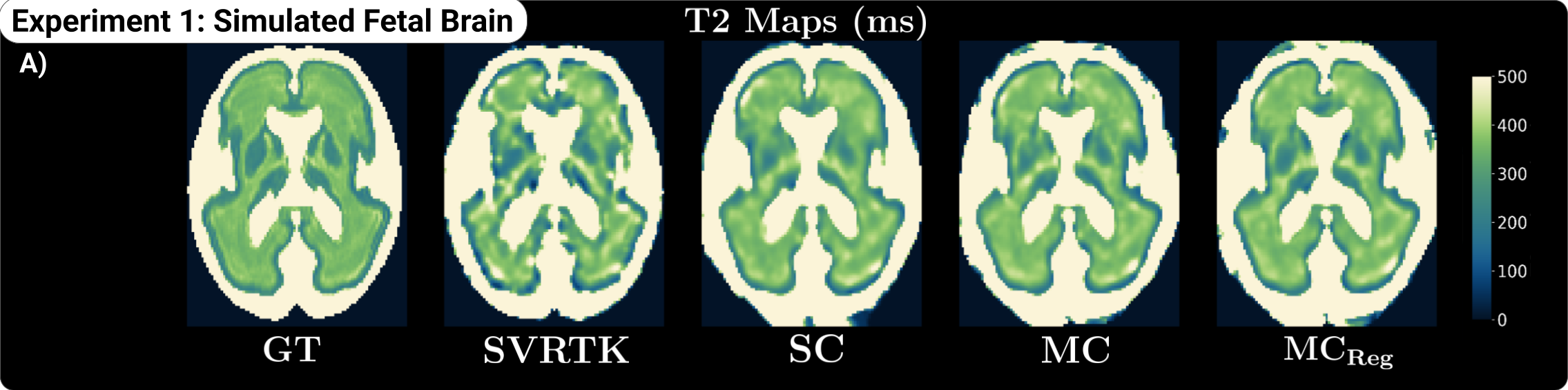}
    \caption{Comparison of T2 maps computed from different SRR methods on FaBian Dataset for GA=27.}
    \label{fig:fabian}
\end{figure}

INR-based methods demonstrate clear superiority over the traditional method SVRTK, both in accurate T2 fitting (MAE) and image quality (SSIM) (see Table~\ref{tab:quant_results}). Figure~\ref{fig:fabian} further indicates that the reconstruction of the GM is particularly challenging for SVRTK, exhibiting severely degraded quality, whereas the INR based methods show consistent T2 fitting.


\subsubsection{Experiment 2 -- Ablation Study on \textit{In Vivo} Adult Data at 0.55T.}


\begin{figure}
\centering
\includegraphics[width=\textwidth]{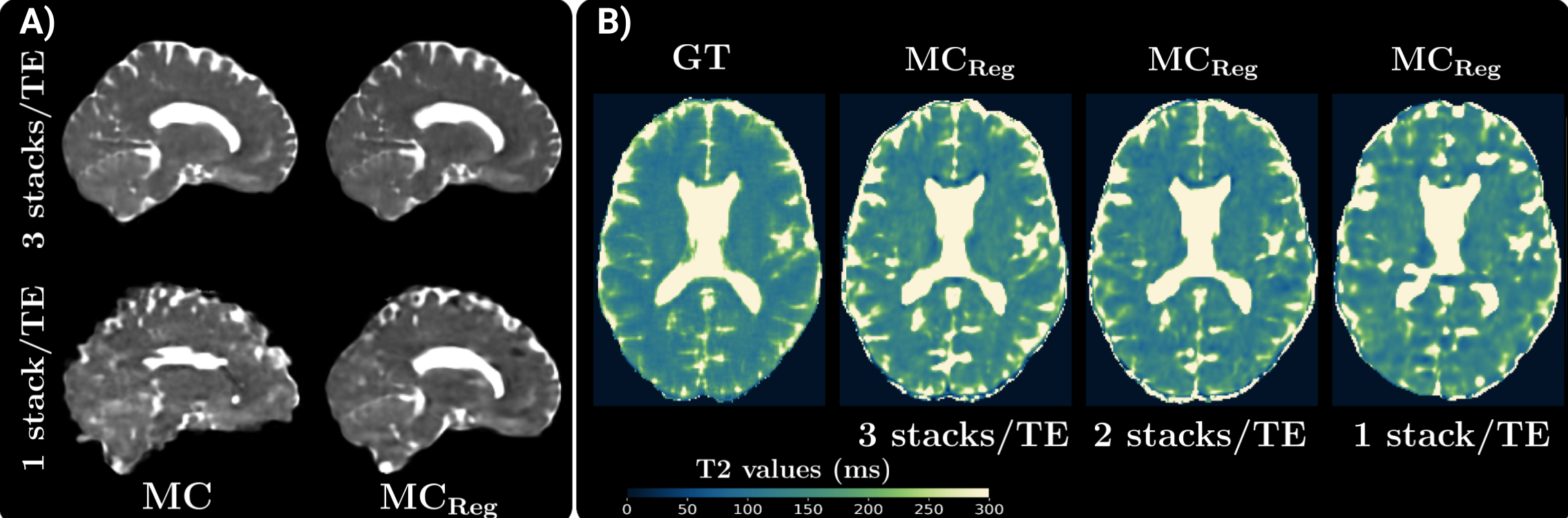}
\vspace{-.1cm}
\captionof{figure}{Visuals for Experiment 2 on Adult Data. A), Comparison of reconstructions for a given subject using either 3 or 1 stack/TE at TE=114ms. B) T2 maps for the same subject obtained using 3, 2 and 1 stack/TE, respectively.
}\label{fig:exp_1_3}


    \centering
    \includegraphics[width=\linewidth]{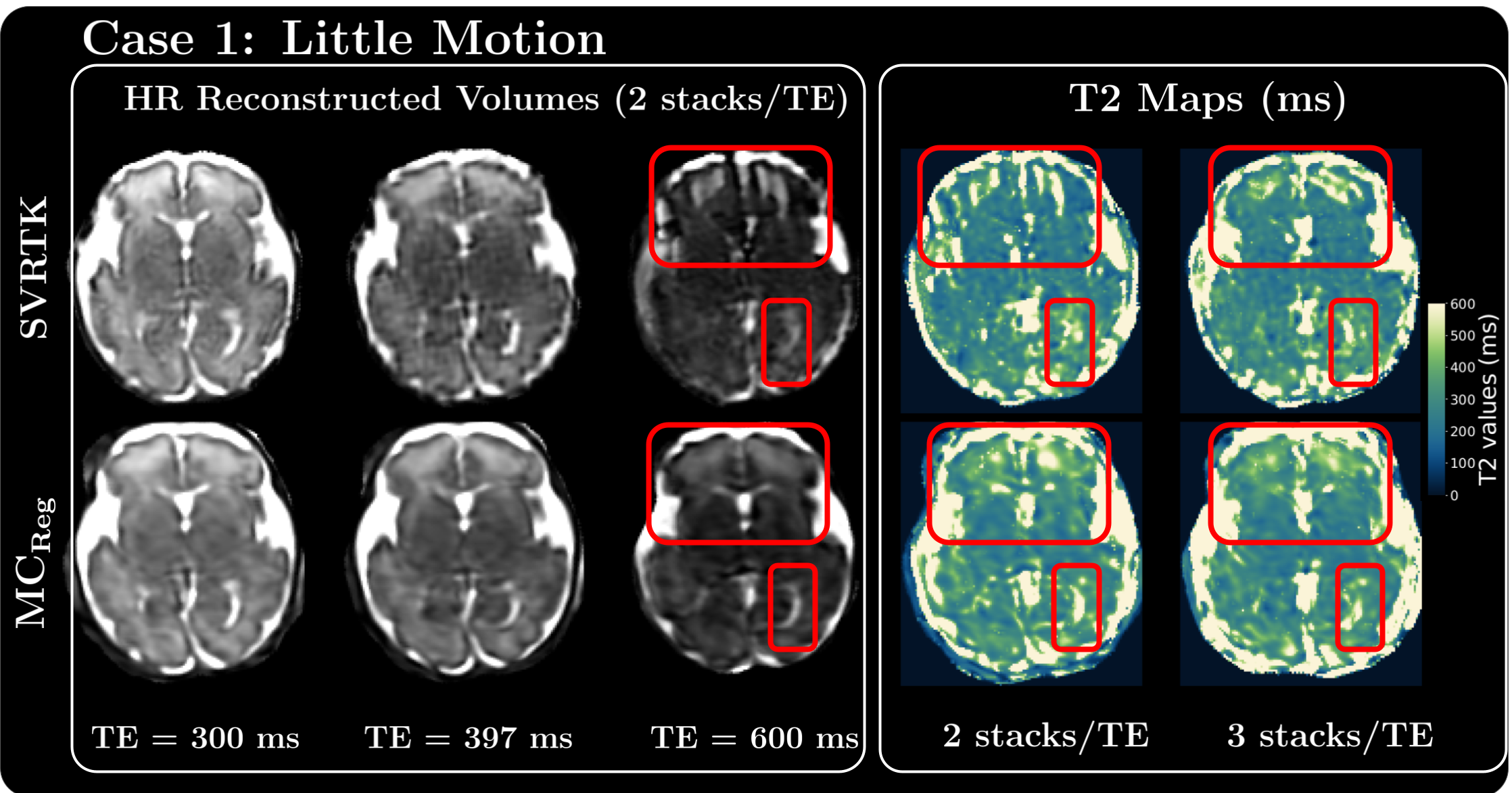}
    \includegraphics[width=\linewidth]{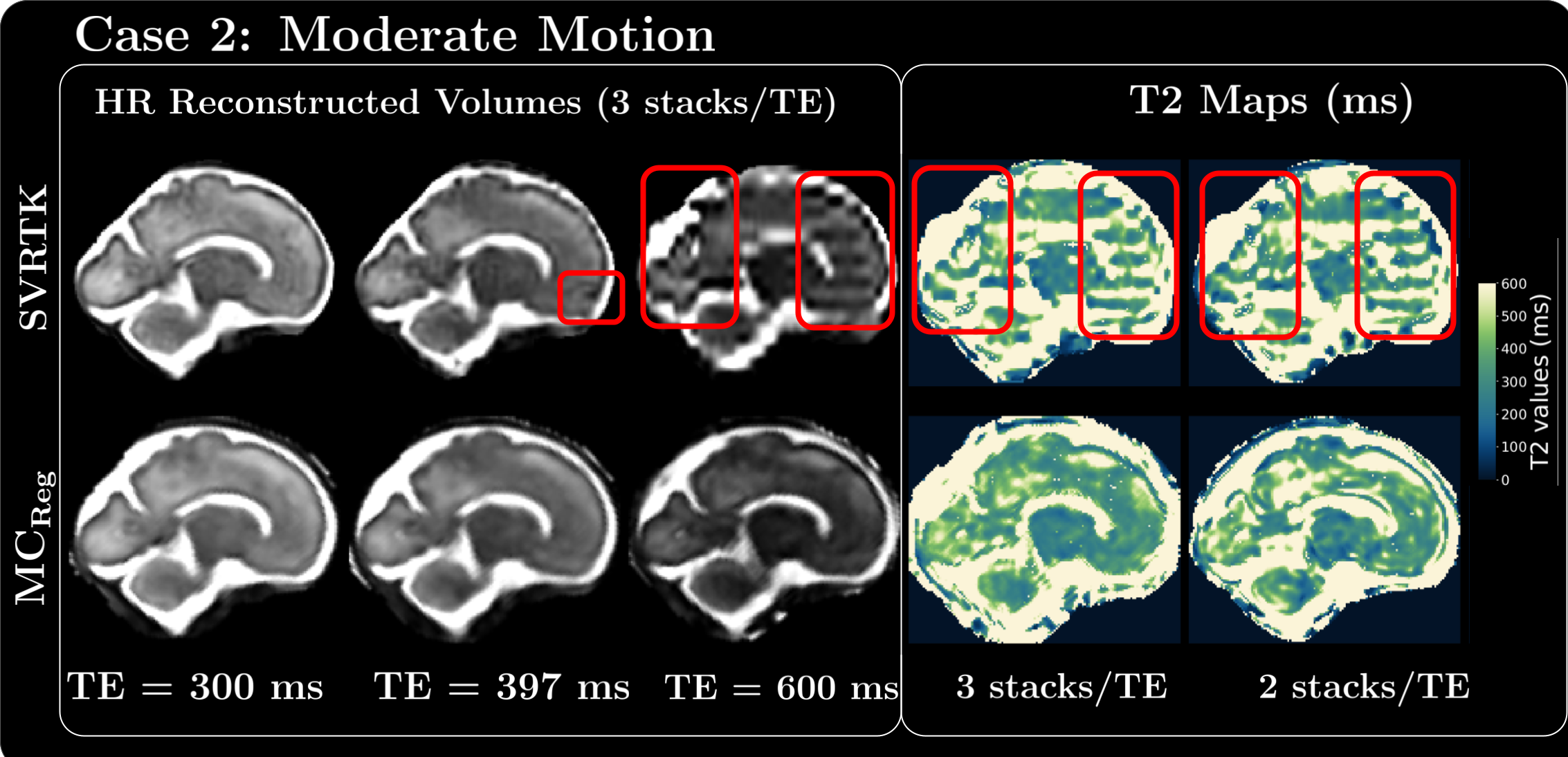} \captionof{figure}{Comparaison of SVRTK and MC$_{\text{Reg}}$ for the reconstruction of the brain and T2 map of the two \textit{in-vivo} fetal subjects with 2 stacks/TE (top) and 3 stacks/TE (bottom).}
    \vspace{-.1cm}
    \label{fig:fetal_invivo}
    \vspace{-.3cm}
\end{figure}

\noindent Quantitative and qualitative results are presented in Table~\ref{tab:adult_data_results} and Fig.~\ref{fig:exp_1_3}, respectively, comparing our proposed framework $\text{MC}_{\mathrm{Reg}}$ over MC and SC \cite{dannecker2025meta} setups. 
For the full input of 3 stacks per TE, we already observe clear improvements of the MC setups over SC\cite{dannecker2025meta}, both in T2 mapping (MAE) and image similarity (SSIM). This difference becomes more striking when decreasing the input to 2 stacks per TE. Under extreme sparsity (one stack per TE), both SC\cite{dannecker2025meta} and MC fail to produce usable reconstructions, In contrast, $\text{MC}_{\mathrm{Reg}}$ still yields reconstructions with SSIM exceeding 0.6, indicating moderate structural similarity, though the MAE remains relatively high.
\begin{table}[h!]
\centering
\scriptsize
\vspace{-0.2cm}
\caption{Region-wise mean and standard deviation of T2 values at 0.55T.}
\label{tab:fetal}
\resizebox{\linewidth}{!}{%
\begin{tabular}{llccccc}
\toprule
\textbf{Subject} & \textbf{Region} & \multicolumn{2}{c}{SVRTK} & \multicolumn{3}{c}{MC$_{\text{Reg}}$} \\
\cmidrule(lr){3-4} \cmidrule(lr){5-7}
& & 3 stacks/TE & 2 stacks/TE & 3 stacks/TE & 2 stacks/TE & 1 stack/TE \\
\midrule
Case 1, GA=32
 & T2 DGM & 248±46 & 245±55 & 246±42 & 251±48 & 250±50 \\
Little Motion & T2 WM & 339±77 & 336±93 & 340±71 & 345±80 & 339±108 \\
\midrule
Case 2, GA=37
 & T2 DGM & 269±95 & 271±103 & 269±41 & 277±61 & 270±72 \\
Moderate Motion & T2 WM & 459±360 & 459±369 & 386±86 & 393±136 & 386±136 \\
\bottomrule
\end{tabular}%
}
\vspace{-0.5cm}
\end{table}

\subsubsection{Experiment 3 -- \textit{In Vivo} Fetal Data at 0.55T.} 
Qualitative results (Fig.~\ref{fig:fetal_invivo}) show that $\text{MC}_{\mathrm{Reg}}$ outperforms SVRTK, especially for TE=600 ms and for under-sparse sampling (2 stacks/TE), maintaining high image quality and anatomical consistency across TEs. In contrast, SVRTK suffers from artefacts, as reflected in the T2 maps depicted in Fig ~\ref{fig:fetal_invivo}. In the low-motion case (case 1), both methods yield similar T2 values in WM (339 ms) and GM (247 ms) (Table ~\ref{tab:fetal}). It suggests robustness in T2 estimation even with imperfect SVRTK reconstructions. However, in the moderate-motion case (case 2), SVRTK overestimates T2 WM by 120 ms with high standard deviations (Table ~\ref{tab:fetal}), while $\text{MC}_{\mathrm{Reg}}$ remains stable. Importantly, $\text{MC}_{\mathrm{Reg}}$ achieves similar T2 estimates even with just 1 stack/TE, though with higher variability. This demonstrates its strong potential for accelerating acquisitions by reducing the number of stacks required per TE.

\section{Conclusion}
Our results show that $\text{MC}_{\mathrm{Reg}}$, by modeling the physics of T2 decay, effectively leverages information across stacks acquired at different TEs to jointly reconstruct HR volumes for all TEs without compromising the accuracy of quantitative T2 maps. Even when maintaining the same input/output ratio as traditional methods (e.g., 9 input stacks for 3 HR outputs), our approach allows more input stacks per TE without requiring acquisitions beyond the standard 3 views, particularly valuable given the already time-consuming nature of T2 mapping. Joint reconstruction also enforces anatomical consistency across TEs, which is often not guaranteed in the presence of severe artifacts. Notably, our findings suggest that fewer than the standard 3 stacks per TE may be sufficient for accurate T2 mapping, thanks to cross-TE information sharing. 
Our method successfully performs T2 mapping at 0.55T with 0.8 mm isotropic resolution, even with reduced acquisition. In addition, the proposed INR framework is more robust to motion than prior methods, which often struggle with high-TE stacks and signal dropouts.
 Future works should explore the selection of the regularization coefficient $\alpha$, which was set empirically in this study and confirm these results on larger \textit{in-vivo} fetal cohort. 
 We believe the proposed T2 mapping framework could enable a more reproducible and robust structural perinatal brain imaging to better depict WM alterations and support multi-centric studies.



\begin{credits}
\subsubsection{\ackname} This research was funded by the Swiss National Science Foundation (215641), the ERC (Deep4MI - 884622), the ANR (AI4CHILD ANR-19-CHIA-0015-01, HINT ANR-22-CE45-0034) and by the ERA-NET NEURON Cofund (MULTI-FACT - 8810003808).
\subsubsection{\discintname}
The authors have no competing interests to declare that are
relevant to the content of this article.
\end{credits}

%
%
%
\newpage
\begingroup
\let\clearpage\relax
\bibliographystyle{ieeetr}
{\small \bibliography{biblio}}
\endgroup

 \newpage
\makeatletter
\setcounter{table}{0}
\setcounter{figure}{0}
\renewcommand 
\thesection{S\@arabic\c@section}
\renewcommand\thetable{S\@arabic\c@table}
\renewcommand\thefigure{S\@arabic\c@figure}
\makeatother

\end{document}